\PassOptionsToPackage{table}{xcolor}
\documentclass[sigconf, screen]{acmart}

\usepackage{booktabs}
\usepackage{makecell}
\usepackage{multirow}
\usepackage{enumitem}
\usepackage{pifont}
\usepackage{float}
\usepackage{tabularx}
\usepackage{balance}

\AtBeginDocument{%
  }

\setcopyright{acmlicensed}

\copyrightyear{2026}
\acmYear{2026}
\setcopyright{cc}
\setcctype{by}
\acmConference[MM '26]{Proceedings of the 34th ACM International Conference on Multimedia}{November 10--14, 2026}{Rio de Janeiro, Brazil}
\acmBooktitle{Proceedings of the 34th ACM International Conference on Multimedia (MM '26), November 10--14, 2026, Rio de Janeiro, Brazil}
\acmDOI{10.1145/3767308.3834753}
\acmISBN{979-8-4007-2213-4/2026/11}

\begin{document}

\title{VCR-Bench: A Modular Open-Source Benchmark for Video Classification Robustness}

\author{Maksim Plinskiy}
\email{maksim.plinskiy@graphics.cs.msu.ru}
\affiliation{%
  \institution{Lomonosov Moscow State University}
  \city{Moscow}
  \country{Russia}
}

\author{Aleksandr Gushchin}
\email{alexander.gushchin@graphics.cs.msu.ru}
\affiliation{%
  \institution{MSU Institute for Artificial Intelligence}
  \city{Moscow}
  \country{Russia}
}

\author{Sergey Lavrushkin}
\email{sergey.lavrushkin@graphics.cs.msu.ru}
\affiliation{%
  \institution{MSU Institute for Artificial Intelligence}
  \city{Moscow}
  \country{Russia}
}

\author{Dmitriy S. Vatolin}
\email{dmitriy@graphics.cs.msu.ru}
\affiliation{%
  \institution{MSU Institute for Artificial Intelligence}
  \city{Moscow}
  \country{Russia}
}

\author{Anastasia Antsiferova}
\email{aantsiferova@graphics.cs.msu.ru}
\affiliation{%
  \institution{MSU Institute for Artificial Intelligence}
  \city{Moscow}
  \country{Russia}
}

\renewcommand{\shortauthors}{Maksim Plinskiy et al.}

\begin{abstract}
Robustness of image classification has several benchmarks, but their video counterparts are absent. In video classification temporal dimension introduces additional degrees of freedom for adversarial attacks, defenses, and preprocessing. Temporal sampling, perturbation budgets, and metric aggregation also interact in ways with no direct analogue in the image setting. Therefore, robustness for video classifiers is studied across scattered, incompatible implementations, making reported numbers hard to reproduce and analyze. We introduce VCR-Bench, a modular open-source benchmark framework that standardizes video loading, wrappers for classifiers, adversarial attacks and defenses, perceptual metrics, configuration presets, and result logging. VCR-Bench currently integrates 30 video classification models, 14 adversarial attacks, and 10 defense wrappers under a common evaluation protocol. We evaluate representative video classifiers, attacks, and defenses on Kinetics-400 subset, reporting clean accuracy, attack success rate, perceptual quality, runtime, and memory usage. VCR-Bench is released with documented installation, reproducible run presets, component-extension interfaces, and scripts for reproducing the reported results at \url{https://github.com/msu-video-group/vcr-bench}.
\end{abstract}



\maketitle

\section{Introduction}

Video classification models are used in application settings where predictions are expected to be reliable across diverse visual conditions, including action recognition, content moderation, sports analytics, medical video analysis, and recommendation systems. Modern convolutional~\cite{tran2015learning,carreira2017quo,feichtenhofer2019slowfast}, transformer-based~\cite{dosovitskiy2021image}, masked-pretraining~\cite{tong2022videomae,wang2023videomaev2}, and multimodal~\cite{wang2021actionclip,radford2021learning} architectures
achieve strong clean accuracy on standard benchmarks such as Kinetics-400~\cite{kay2017kinetics}. However, clean accuracy measures performance only on unperturbed inputs. It does not answer whether a classifier remains reliable when the input is deliberately manipulated. In adversarial attacks, a small, often visually inconspicuous perturbation is added to the input in order to change the model prediction~\cite{goodfellow2015explaining,kurakin2017adversarialscale}. This threat model is practically relevant for video systems: for example, an attacker may seek to evade content moderation, alter the recognized action in a surveillance or sports-analysis pipeline, or suppress a safety-critical prediction while keeping the video perceptually plausible to a human viewer. Robustness evaluation is essential for reliable performance.

For image classification, adversarial robustness has led to mature evaluation infrastructure, including standardized attack libraries, benchmarks, and reproducible leaderboards~\cite{papernot2016technical,rauber2017foolbox,croce2020reliable,croce2021robustbench}. Video classification lacks comparable infrastructure. Its robustness evaluation depends on additional protocol choices---clip length, frame sampling, temporal stride, spatial preprocessing, normalization, perturbation budget, clip-to-video aggregation, and whether attacks optimize frame-level, temporal, or video-level predictions. Defenses further complicate evaluation, since preprocessing, stochasticity, temporal smoothing, or purification can make non-adaptive attacks overestimate robustness. Consequently, reported robustness can reflect implementation and protocol choices as much as genuine model behavior.

VCR-Bench addresses this gap with a unified open-source benchmark for adversarial robustness evaluation in video classification. It standardizes video loading, model wrappers, attack and defense interfaces, preprocessing, temporal sampling, perceptual metrics, resource logging, and reproducible configuration presets. To the best of our knowledge, VCR-Bench is the first open-source benchmark that jointly supports clean evaluation, adversarial attacks, and defenses for video classification under a common protocol, enabling models, attacks, and defenses to be compared as components of the same evaluation pipeline rather than as isolated results produced under incompatible assumptions.

\begin{figure}[t]
  \centering
  \begin{minipage}[c]{0.92\linewidth}
      \centering
      \includegraphics[width=\linewidth]{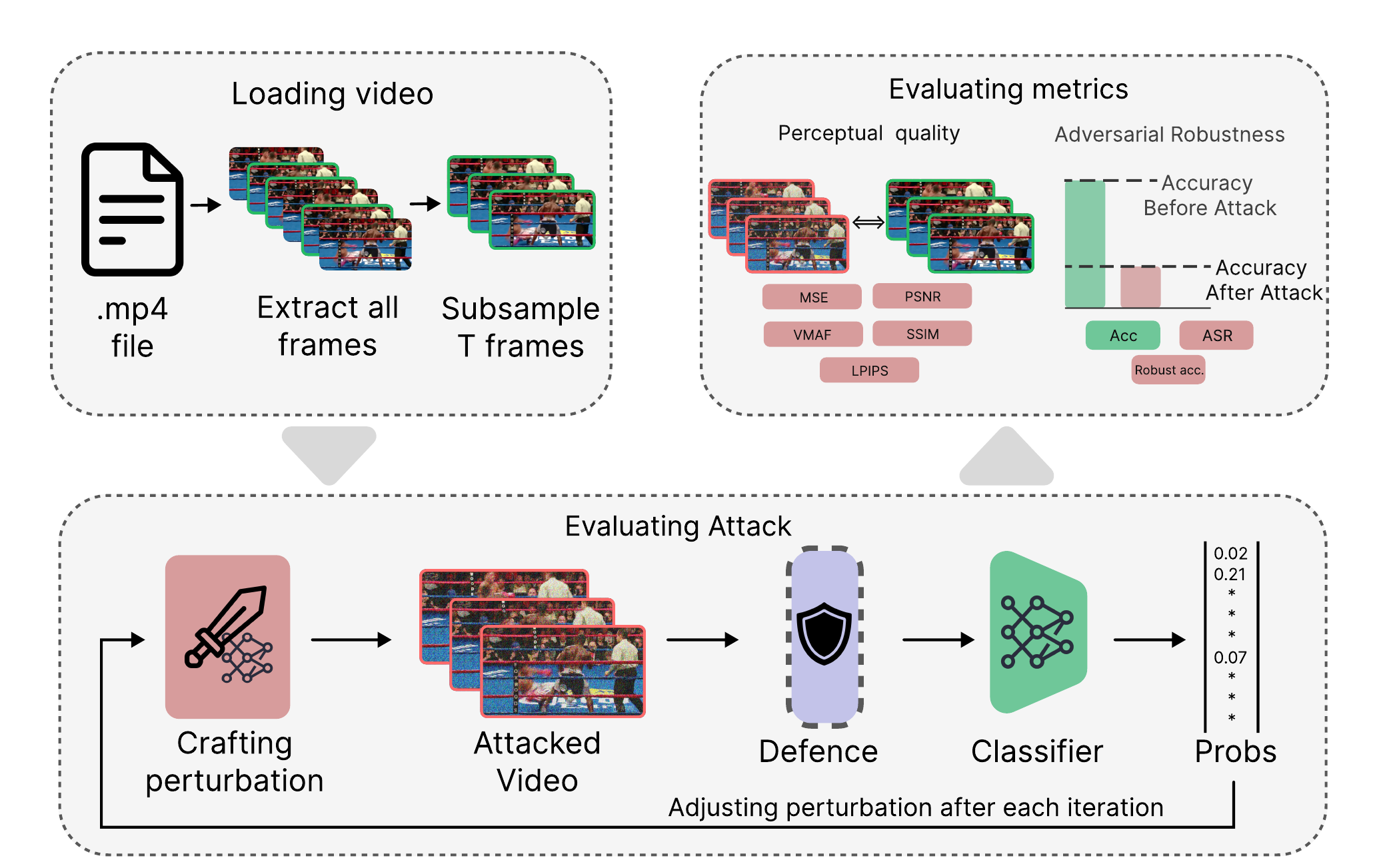}
    \end{minipage}%
  \caption{VCR-Bench evaluation pipeline.}
  \Description{Scheme}
  \label{fig:pipeline}
  \vspace{-10pt}
\end{figure}

Our contributions are as follows:
\begin{itemize}[leftmargin=*]
  \item We release VCR-Bench, an open-source benchmark framework that integrates datasets, video classifiers, adversarial attacks, defenses, perceptual metrics, configuration presets, and result logging under one evaluation protocol.
  \item We design a modular architecture with stable component interfaces and convention-based discovery, allowing new models, attacks, defenses, and datasets to be added without rewriting the core evaluation loop.
  \item We provide standardized experiment configurations for clean evaluation, adversarial evaluation, covering clip sampling, preprocessing, perturbation budgets, attack success, clean / defended accuracy, perceptual distortion, runtime, and memory usage.
  \item We include 30 video classifiers, 14 adversarial attacks, and 10 defense methods, together with documented run presets, exact command templates, output schemas, logging utilities, and extension examples for reproducible benchmarking and extension.
\end{itemize}

\section{VCR-Bench Framework}

\subsection{Evaluation Pipeline}

Figure~\ref{fig:pipeline} shows the VCR-Bench evaluation pipeline. A raw video is decoded into a tensor representation and sampled into clips using a deterministic policy for reproducibility. The sampled clips can optionally be passed through a defensive transformation to measure its impact on clean accuracy before model-specific spatial resizing, cropping, normalization, and tensor-layout conversion. The classifier then produces logits for each clip, and the framework aggregates predictions into a per-video label. During adversarial evaluation, an attack module iteratively produces adversarial perturbations and queries the defended classifier until it reaches a stopping condition or exhausts its budget. If attack is non-adaptive, attack queries original classifier without applying defense. Only the final evaluation is made with the defended model.

The pipeline records per-video outputs, including clean prediction, attacked prediction, attack success rate, perceptual distortion metrics, runtime, and memory statistics. 

\subsection{Models, Attacks, and Defenses}

Table~\ref{tab:components} summarizes 
the current coverage. VCR-Bench is organized around component families 
rather than a fixed experiment script. The model registry currently 
covers 30 video classification models spanning convolutional 
architectures such as C3D~\cite{tran2015learning} and 
I3D~\cite{carreira2017quo}; 
temporal-shift networks including TSN~\cite{wang2016temporal} and 
TSM~\cite{lin2019tsm}; 
3D-convolutional models such as R(2+1)D~\cite{tran2018closer} and 
X3D~\cite{feichtenhofer2020x3d}; 
transformer-based models including 
TimeSformer~\cite{bertasius2021space}, 
MViTv2~\cite{li2022mvitv2}, and 
Video Swin~\cite{liu2022video}; 
masked-pretraining models such as
VideoMAEv2~\cite{wang2023videomaev2} and 
UMT~\cite{li2023unmasked}; 
and multimodal models such as 
ActionCLIP~\cite{wang2021actionclip}, 
InternVideo2~\cite{wang2024internvideo2}, and 
ONE-PEACE~\cite{wang2024onepeace}.

The attack registry currently covers 14 methods, including 
gradient-based white-box attacks such as 
I-FGSM~\cite{kurakin2017adversarialscale}, 
MI-FGSM~\cite{dong2018boosting}, perceptual attacks including 
SSAH~\cite{luo2022frequency} and StAdv~\cite{xiao2018spatially}; 
query-based black-box attacks such as 
Gradient Estimation Attack~\cite{li2021adversarial}; 
universal perturbations (UAP)~\cite{moosavi2017universal}; 
and video-specific attacks including 
StyleFool~\cite{cao2023stylefool}, and 
BMTC~\cite{li2025bmtc}.

The defense registry currently covers 10 purification methods, 
including simple input transformations such as 
JPEG compression and crop-resize~\cite{guo2018countering}, 
input randomization~\cite{xie2018mitigating}, and 
randomized smoothing~\cite{cohen2019certified}; 
and video-specific transformations such as 
temporal shuffling~\cite{hwang2024temporal} and 
diffusion-based purification (VideoPure)~\cite{jiang2025videopure}.

\begin{table*}[t]
  \centering
  \small
  \setlength{\tabcolsep}{4pt}
  \caption{VCR-Bench provides systematic coverage across core components: diverse model families, multiple attack paradigms, and defense methods spanning spatial, temporal, and diffusion-based purification.}
  \label{tab:components}
  \begin{tabular}{p{0.12\linewidth}p{0.05\linewidth}p{0.78\linewidth}}
    \toprule
    Component & Count & Implemented \\
    \midrule
    Models & 30 & 
      C3D~\cite{tran2015learning},
      TSN~\cite{wang2016temporal},
      I3D~\cite{carreira2017quo},
      C2D / Non-local~\cite{wang2018nonlocal},
      R(2+1)D~\cite{tran2018closer},
      SlowOnly~\cite{hara2018can},
      TRN~\cite{zhou2018temporal},
      TSM~\cite{lin2019tsm},
      SlowFast~\cite{feichtenhofer2019slowfast},
      CSN~\cite{tran2019video},
      TIN~\cite{shao2020temporal},
      TPN~\cite{yang2020temporal},
      X3D~\cite{feichtenhofer2020x3d},
      TANet~\cite{liu2021tam},
      TimeSformer~\cite{bertasius2021space},
      ViViT~\cite{arnab2021vivit},
      MViT~\cite{fan2021multiscale},
      Video Swin~\cite{liu2022video},
      MViTv2~\cite{li2022mvitv2},
      UniFormer~\cite{li2022uniformer},
      VideoMAE~\cite{tong2022videomae},
      InternVideo~\cite{wang2022internvideo},
      UniFormerV2~\cite{li2023uniformerv2},
      UMT~\cite{li2023unmasked},
      VideoMAEv2~\cite{wang2023videomaev2},
      TAdaFormer~\cite{huang2023temporally},
      AMD~\cite{zhao2024asymmetric},
      InternVideo2~\cite{wang2024internvideo2},
      ONE-PEACE~\cite{wang2024onepeace},
      ActionCLIP~\cite{wang2021actionclip},
      ILA~\cite{tu2023ila} \\
    \midrule
    Attacks & 14 & 
      I-FGSM~\cite{kurakin2017adversarialscale},
      MI-FGSM~\cite{dong2018boosting},
      AMI-FGSM ,
      UAP~\cite{moosavi2017universal},
      StAdv~\cite{xiao2018spatially},
      Square Attack~\cite{andriushchenko2020square},
      GradEst~\cite{li2021adversarial},
      Korhonen et al.~\cite{korhonen2022adversarial},
      Zhang et al. (3 variants)~\cite{zhang2022perceptual},
      SSAH~\cite{luo2022frequency},
      StyleFool~\cite{cao2023stylefool},
      BMTC~\cite{li2025bmtc} \\
    \midrule
    Defenses & 10 & 
      JPEG compression~\cite{guo2018countering},
      Crop-resize~\cite{guo2018countering},
      Randomized smoothing~\cite{cohen2019certified},
      Temporal shuffling~\cite{hwang2024temporal},
      VideoPure~\cite{jiang2025videopure},
      FreqPure~\cite{pei2025freqpure},
      Gaussian blur,
      Temporal median,
      Flip/rotate \\
    \midrule
    Metrics & 10 & 
      Clean / robust / defended accuracy, 
      Attack success rate (ASR), 
      MSE, PSNR, 
      SSIM~\cite{wang2004image}, 
      LPIPS~\cite{zhang2018unreasonable}, 
      DISTS~\cite{ding2022dists}, 
      VMAF~\cite{li2018vmaf}, 
      Runtime (s/video), 
      Peak GPU memory \\
    \bottomrule
  \end{tabular}
\end{table*}

\subsection{Metrics, Outputs, and Reproducibility}

VCR-Bench reports task performance, perceptual distortion, and computational cost under a common logging format. Task-level metrics include clean accuracy, robust accuracy, defended clean accuracy, defended robust accuracy, and attack success rate (ASR). Clean accuracy is measured on unperturbed videos, while robust accuracy is measured after applying an adversarial attack. For defenses, defended clean accuracy measures the effect of the defense on clean inputs, and defended robust accuracy measures classification performance after both attack and defense. The Attack Success Rate is computed over videos that are correctly classified by the clean model, $ASR = N_{\mathrm{success}} / N_{\mathrm{clean\ correct}}$. Perceptual and distortion metrics include MSE, PSNR, SSIM~\cite{wang2004image}, LPIPS~\cite{zhang2018unreasonable}, and VMAF~\cite{li2018vmaf}. 

The framework writes structured outputs as CSV and JSON files. Every run records the dataset subset, model preset, attack preset, defense preset, random seed, clip-sampling settings, perturbation budget, metric values, and execution metadata. VCR-Bench also supports qualitative analysis by exporting clean, adversarial, and defended videos for side-by-side visual inspection.

\subsection{Configuration, Extensibility, and Release}
VCR-Bench builds on existing video-understanding toolchains such as MMAction2~\cite{2020mmaction2} and PyTorchVideo~\cite{fan2021pytorchvideo}, but adds a standardized adversarial-robustness evaluation on top. The framework standardizes how videos are loaded, models are evaluated, attacks and defenses are applied, and results are recorded.

Experiments are specified by combining four types of configuration: shared defaults, local path settings, reusable component presets, and run presets. Defaults define common benchmark behavior; local settings keep machine-specific paths out of experiment files; component presets describe individual datasets, models, attacks, defenses, and metrics; and each run preset defines one complete experiment.

The extension interface is based on stable component abstractions. New classifiers implement \texttt{BaseVideoClassifier}, new attacks implement \texttt{BaseVideoAttack}, and new defenses implement \texttt{BaseDefense}. Component discovery follows a convention-based layout: a new component is placed under the corresponding package namespace and exposed through its preset. For attacks that require precomputation, such as BMTC, VCR-Bench provides an offline preparation hook. 

The framework can run from local presets on a single machine, with optional Slurm helpers for larger sweeps. The release includes source code, installation instructions, quick-start examples, component documentation, and presets for the experiments in this paper. The repository is available at \url{https://github.com/msu-video-group/vcr-bench}, and the leaderboard is hosted at \url{https://msu-video-group.github.io/vcr-bench/}. VCR-Bench is released under the MIT license; datasets, pretrained checkpoints, and third-party components retain their original licenses and download terms.

\section{Evaluation and Results}

\subsection{Experimental Setup}

We evaluate VCR-Bench on Kinetics-400 test subset. The main experiments use a fixed subset of 100 videos selected randomly. Kinetics-400 is used because it is the standard large-scale action-recognition benchmark supported by most integrated classifiers, allowing heterogeneous architectures to be evaluated under a common label space and data protocol. Unless otherwise stated, all runs use deterministic temporal sampling with seed 42, and model-specific preprocessing supplied by each model wrapper. Experiments are run on NVIDIA A100 80GB with this software environment: Python 3.10, PyTorch 2.7.1 built with CUDA 11.8, cuDNN 9.1, torchvision, OpenCV, scikit-image, and FFmpeg/libvmaf for video-quality metrics.

For adversarial evaluation, we use a perturbation budget of  $L_\infty \epsilon=8/255$ for white-box attacks, 20 and 200 iterations for iterative white and black box attacks respectively as these parameter values correspond to standard values across related works in adversarial machine learning~\cite{madry2017towards}. Defense experiments report both clean-performance drop and defended robustness. When a defense is differentiable or can be approximated, we report adaptive evaluation results; otherwise, we mark the setting as non-adaptive and discuss the limitation.

\begin{table*}[t]
  \centering
  \small
  \setlength{\tabcolsep}{3pt}
  \caption{
  Benchmark summary averaged by model across all attack launches under adaptive targeted attacks. Time/video reports the mean processing time per video in seconds. Best results are shown in bold, and second-best results are underlined.
  }
  \label{tab:main_results_by_model}
  \begin{tabularx}{\textwidth}{lXXXXXXXXXX}
    \toprule
    Model & Clean acc. & ASR & Robust acc. & MSE & PSNR & SSIM & LPIPS & VMAF & Iters & Time/video \\
    \midrule
    ActionCLIP & 66.91 & 72.38 & 18.48 & 8.53 & 43.88 & 0.9496 & 0.0539 & 96.61 & 20.00 & 6.64 \\
    AMD & 77.75 & 72.55 & 21.34 & 7.51 & 44.83 & 0.9542 & 0.0582 & 94.04 & \textbf{10.96} & 8.95 \\
    C2D & 66.93 & 69.21 & 20.61 & 10.36 & 43.71 & 0.9347 & 0.0749 & \underline{97.44} & 12.86 & 10.46 \\
    ILA & 79.64 & 66.17 & 26.95 & 14.38 & 42.52 & 0.9123 & 0.0841 & 95.39 & 16.24 & 5.99 \\
    MViTv2 & 75.66 & 67.57 & 24.54 & \textbf{6.04} & \textbf{46.33} & \textbf{0.9626} & \underline{0.0487} & 96.76 & 11.98 & 21.12 \\
    ONE-PEACE & 71.09 & 60.66 & 27.96 & \underline{7.08} & 45.47 & 0.9536 & \textbf{0.0349} & 96.98 & 14.64 & 35.74 \\
    R(2+1)D & 46.96 & 76.92 & 10.84 & 12.98 & 42.85 & 0.9180 & 0.0985 & 95.96 & 15.66 & 13.67 \\
    TAdaFormer & 79.96 & \textbf{49.50} & \textbf{40.38} & 15.68 & 41.94 & 0.9167 & 0.0899 & 92.48 & 18.86 & 31.22 \\
    TANet & 60.55 & 67.03 & 19.96 & 13.04 & 42.68 & 0.9212 & 0.0998 & 96.18 & 15.21 & 15.57 \\
    TimeSformer & 75.18 & 66.06 & 25.52 & 10.79 & 42.90 & 0.9410 & 0.0754 & 90.92 & 13.30 & \textbf{2.26} \\
    TPN & 63.12 & 59.63 & 25.48 & 9.33 & 44.33 & 0.9432 & 0.0756 & 96.96 & 13.38 & 12.96 \\
    TSM & 63.09 & \underline{54.91} & 28.45 & 9.89 & 43.53 & 0.9422 & 0.0781 & 95.17 & 14.72 & 7.15 \\
    TSMNonLocal & 62.64 & 64.22 & 22.41 & 12.38 & 42.57 & 0.9279 & 0.0913 & 93.25 & 14.38 & 5.11 \\
    TSN & 67.84 & 58.41 & 28.21 & 7.25 & \underline{45.96} & \underline{0.9567} & \underline{0.0487} & 96.77 & 16.22 & \underline{4.25} \\
    UniFormer & 76.51 & 68.37 & 24.20 & 7.22 & 45.79 & 0.9544 & 0.0512 & \textbf{97.52} & 13.47 & 17.69 \\
    UniFormerV2 & \textbf{86.73} & 68.89 & 26.98 & 10.40 & 43.62 & 0.9373 & 0.0668 & 93.95 & 12.53 & 7.08 \\
    Video Swin & 74.77 & 62.03 & 28.39 & 11.71 & 43.85 & 0.9275 & 0.0704 & 96.29 & 13.52 & 31.55 \\
    VideoMAE & 76.70 & 69.06 & 23.73 & 8.23 & 44.41 & 0.9506 & 0.0630 & 96.92 & \underline{11.37} & 13.09 \\
    VideoMAEv2 & \underline{84.12} & 66.02 & \underline{28.59} & 10.09 & 43.98 & 0.9399 & 0.0731 & 96.06 & 12.69 & 14.24 \\
    \bottomrule
  \end{tabularx}
\end{table*}

\begin{table}[t]
  \centering
  \tiny
  \setlength{\tabcolsep}{1.4pt}
  \renewcommand{\arraystretch}{0.92}
  \caption{Defence-wise robustness by attack. Each cell reports defended robust accuracy / ASR (\%), aggregated across all models. Bold indicates the best robust accuracy in each column; underline indicates the second best.}
  \label{tab:defence_attack_matrix_inline}
  \resizebox{\columnwidth}{!}{%
  \begin{tabular}{lccccccc}
    \toprule
    Defence & I-FGSM & MI-FGSM & AMI-FGSM & Korhonen & ST-Adv & \makecell[c]{Zhang\\DISTS} & \makecell[c]{Zhang\\LPIPS} \\
    \midrule
    No defence    & 5.2/93.0  & 7.8/89.5  & 8.0/89.2  & 9.2/87.5  & 67.0/9.6  & 65.0/12.6 & 3.8/93.0 \\
    Crop-resize   & \underline{9.2/87.5}  & \underline{16.1/78.1} & \underline{15.0/79.7} & \underline{56.0/24.5} & \textbf{71.4/3.9}  & \textbf{71.0/4.0}  & \underline{6.8/87.7} \\
    Flip          & 5.5/92.6  & 8.5/88.5  & 8.3/88.8  & 10.9/85.2 & 66.6/10.1 & 63.5/14.3 & 4.1/92.6 \\
    Gaussian blur & 5.2/92.5  & 9.5/86.0  & 9.6/85.9  & 34.7/49.1 & 61.8/8.6  & 61.1/10.0 & 4.4/91.0 \\
    Rotate        & \textbf{14.0/79.4} & \textbf{22.5/67.5} & \textbf{23.3/66.7} & \textbf{58.0/16.8} & 63.5/7.7  & 64.2/7.6  & \textbf{8.0/84.2} \\
    Shuffle       & 5.3/92.8  & 8.8/88.0  & 9.1/87.8  & 24.9/66.3 & \underline{67.2/8.8}  & \underline{68.7/7.1}  & 3.6/93.3 \\
    Temp. median  & 3.8/94.6  & 4.8/93.1  & 5.0/93.0  & 8.7/87.8  & 61.8/12.3 & 60.5/14.0 & 2.9/94.5 \\
    \bottomrule
  \end{tabular}%
  }
\end{table}

\subsection{Results}

\paragraph{Clean Model Evaluation.}

Table~\ref{tab:main_results_by_model} presents the aggregated results for videoclassification models averaged over all attack methods. 
Overall, the results show that adaptive targeted attacks remain highly effective against most video classification architectures, with many models suffering ASR values above 60\% despite strong clean accuracy. Robustness is therefore not explained by clean performance alone: UniFormerV2 achieves the highest clean accuracy, yet remains highly vulnerable, while older temporal aggregation models such as TSM, TSN, and TPN obtain more competitive ASR despite weaker clean accuracy. The clearest outlier is TAdaFormer, which achieves the lowest ASR and highest robust accuracy, suggesting that temporally adaptive aggregation makes targeted manipulation less reliable. However, this robustness comes with substantially larger perceptual distortion, indicating that successful attacks against TAdaFormer require less stealthy perturbations. In contrast, models such as MViTv2, ONE-PEACE, TSN, and UniFormer allow attacks with high visual fidelity. These trends suggest that robustness is governed less by broad architectural categories such as CNNs versus Transformers, and more by how temporal evidence is aggregated across frames. Structured temporal mechanisms, temporal adaptivity, and constrained aggregation appear more relevant to adversarial resistance than model scale, clean accuracy, or generic pretraining alone.

\paragraph{Defense Evaluation.}
Table \ref{tab:defence_attack_matrix_inline} reports the effectiveness of defense methods. Defense experiments evaluate whether preprocessing or purification wrappers improve robustness without destroying clean performance. For each defense, VCR-Bench reports defended clean accuracy, defended robust accuracy, post-defense ASR, perceptual distortion, runtime overhead, and memory use. The purpose is to identify defenses that remain useful under the strongest applicable evaluation protocol. Defense comparison should therefore be read primarily as a search for the best accuracy--robustness trade-off.

\paragraph{Runtime, Perceptual Quality, and Reproducibility.}
The benchmark records perceptual distortion and computational cost because attack success alone is not enough to compare attacks fairly. However, these quantities serve different users. For model selection, robust accuracy and ASR are primary: the user wants to know which classifier fails least often under attack. Runtime, memory, and query budget are secondary diagnostic variables: they explain whether an attack is practically usable and whether a robustness result may reflect under-optimization rather than genuine resistance. Detailed per-attack results on the benchmark webpage identify which attack families dominate under fixed perturbation and query budgets; Table~\ref{tab:main_results_by_model} instead provides the compact model-level view.

\section{Limitations}

VCR-Bench is intended as a standardized benchmark framework, not a proof that any particular video classifier is secure. The current evaluation is limited to Kinetics-400 subset and may not transfer to all video domains, data distributions, or deployment settings. Addressing this, we extend our codebase with support of UCF101~\cite{soomro2012ucf101} and Something-Something~v2~\cite{goyal2017something}. The perturbation budget and attack settings represent a specific operating point; stronger budgets, longer attacks, or adaptive attack variants may change method rankings.
The breadth of the framework also creates engineering constraints. Some model, attack, and defense combinations may be incompatible. 

\section{Conclusion}

We introduced VCR-Bench, a modular open-source benchmark framework for evaluating adversarial robustness in video classification. The framework defines a common experimental contract for datasets, video classifiers, adversarial attacks, defense methods, perceptual metrics, configuration presets, and structured result logs. Our representative evaluation demonstrates how this contract supports joint reporting of clean accuracy, attacked and defended performance, perceptual distortion, runtime, and memory usage, while reproducible presets and recorded execution metadata make the resulting comparisons auditable rather than tied to one-off scripts. The released code, documentation, run presets, output schemas, extension interfaces, and leaderboard provide a reusable baseline for comparing existing methods and integrating future models, attacks, defenses, and datasets under the same protocol. 

\begin{acks}
The research was carried out using the MSU-270 supercomputer of Lomonosov Moscow State University.
\end{acks}
\clearpage

\bibliographystyle{ACM-Reference-Format}
\balance
\bibliography{sample-base}

@String{Computer = "{IEEE} Computer" }

@String{Springer = "Springer-Verlag" }

@inproceedings{feichtenhofer2019slowfast,
  title={Slowfast networks for video recognition},
  author={Feichtenhofer, Christoph and Fan, Haoqi and Malik, Jitendra and He, Kaiming},
  booktitle={Proceedings of the IEEE/CVF international conference on computer vision},
  pages={6202--6211},
  year={2019}
}

@article{li2018vmaf,
  title={VMAF: The journey continues},
  author={Li, Zhi and Bampis, Christos and Novak, Julie and Aaron, Anne and Swanson, Kyle and Moorthy, Anush and Cock, JD},
  journal={Netflix Technology Blog},
  volume={25},
  number={1},
  year={2018}
}

@article{madry2017towards,
  title={Towards deep learning models resistant to adversarial attacks},
  author={Madry, Aleksander and Makelov, Aleksandar and Schmidt, Ludwig and Tsipras, Dimitris and Vladu, Adrian},
  journal={arXiv preprint arXiv:1706.06083},
  year={2017}
}

@inproceedings{goodfellow2015explaining,
  title={Explaining and Harnessing Adversarial Examples},
  author={Goodfellow, Ian J. and Shlens, Jonathon and Szegedy, Christian},
  booktitle={International Conference on Learning Representations (ICLR)},
  year={2015}
}

@inproceedings{tran2015learning,
  title={Learning Spatiotemporal Features with {3D} Convolutional Networks},
  author={Tran, Du and Bourdev, Lubomir and Fergus, Rob and Torresani, Lorenzo and Paluri, Manohar},
  booktitle={Proceedings of the IEEE International Conference on Computer Vision (ICCV)},
  pages={4489--4497},
  year={2015},
  doi={10.1109/ICCV.2015.510}
}

@inproceedings{wang2016temporal,
  title={Temporal Segment Networks: Towards Good Practices for Deep Action Recognition},
  author={Wang, Limin and Xiong, Yuanjun and Wang, Zhe and Qiao, Yu and Lin, Dahua and Tang, Xiaoou and Van Gool, Luc},
  booktitle={Proceedings of the European Conference on Computer Vision (ECCV)},
  pages={20--36},
  year={2016},
  publisher={Springer}
}

@inproceedings{carreira2017quo,
  title={Quo Vadis, Action Recognition? {A} New Model and the Kinetics Dataset},
  author={Carreira, Joao and Zisserman, Andrew},
  booktitle={Proceedings of the IEEE Conference on Computer Vision and Pattern Recognition (CVPR)},
  pages={6299--6308},
  year={2017},
  doi={10.1109/CVPR.2017.502}
}

@inproceedings{wang2018nonlocal,
  title={Non-local Neural Networks},
  author={Wang, Xiaolong and Girshick, Ross and Gupta, Abhinav and He, Kaiming},
  booktitle={Proceedings of the IEEE/CVF Conference on Computer Vision and Pattern Recognition (CVPR)},
  pages={7794--7803},
  year={2018},
  doi={10.1109/CVPR.2018.00813}
}

@inproceedings{tran2018closer,
  title={A Closer Look at Spatiotemporal Convolutions for Action Recognition},
  author={Tran, Du and Wang, Heng and Torresani, Lorenzo and Ray, Jamie and LeCun, Yann and Paluri, Manohar},
  booktitle={Proceedings of the IEEE/CVF Conference on Computer Vision and Pattern Recognition (CVPR)},
  pages={6450--6459},
  year={2018},
  doi={10.1109/CVPR.2018.00675}
}

@inproceedings{hara2018can,
  title={Can Spatiotemporal {3D} {CNNs} Retrace the History of {2D} {CNNs} and {ImageNet}?},
  author={Hara, Kensho and Kataoka, Hirokatsu and Satoh, Yutaka},
  booktitle={Proceedings of the IEEE/CVF Conference on Computer Vision and Pattern Recognition (CVPR)},
  pages={6546--6555},
  year={2018},
  doi={10.1109/CVPR.2018.00685}
}

@inproceedings{zhou2018temporal,
  title={Temporal Relational Reasoning in Videos},
  author={Zhou, Bolei and Andonian, Alex and Oliva, Aude and Torralba, Antonio},
  booktitle={Proceedings of the European Conference on Computer Vision (ECCV)},
  pages={803--818},
  year={2018},
  publisher={Springer}
}

@inproceedings{lin2019tsm,
  title={{TSM}: Temporal Shift Module for Efficient Video Understanding},
  author={Lin, Ji and Gan, Chuang and Han, Song},
  booktitle={Proceedings of the IEEE/CVF International Conference on Computer Vision (ICCV)},
  pages={7083--7093},
  year={2019},
  doi={10.1109/ICCV.2019.00718}
}

@inproceedings{tran2019video,
  title={Video Classification with Channel-Separated Convolutional Networks},
  author={Tran, Du and Wang, Heng and Torresani, Lorenzo and Feiszli, Matt},
  booktitle={Proceedings of the IEEE/CVF International Conference on Computer Vision (ICCV)},
  pages={5552--5561},
  year={2019},
  doi={10.1109/ICCV.2019.00565}
}

@inproceedings{shao2020temporal,
  title={Temporal Interlacing Network},
  author={Shao, Hao and Qian, Shengju and Liu, Yu},
  booktitle={Proceedings of the AAAI Conference on Artificial Intelligence},
  volume={34},
  number={7},
  pages={11966--11973},
  year={2020},
  doi={10.1609/aaai.v34i07.6872}
}

@inproceedings{yang2020temporal,
  title={Temporal pyramid network for action recognition},
  author={Yang, Ceyuan and Xu, Yinghao and Shi, Jianping and Dai, Bo and Zhou, Bolei},
  booktitle={Proceedings of the IEEE/CVF conference on computer vision and pattern recognition},
  pages={591--600},
  year={2020}
}

@inproceedings{feichtenhofer2020x3d,
  title={{X3D}: Expanding Architectures for Efficient Video Recognition},
  author={Feichtenhofer, Christoph},
  booktitle={Proceedings of the IEEE/CVF Conference on Computer Vision and Pattern Recognition (CVPR)},
  pages={203--213},
  year={2020},
  doi={10.1109/CVPR42600.2020.00028}
}

@inproceedings{bertasius2021space,
  title={Is Space-Time Attention All You Need for Video Understanding?},
  author={Bertasius, Gedas and Wang, Heng and Torresani, Lorenzo},
  booktitle={Proceedings of the 38th International Conference on Machine Learning (ICML)},
  pages={813--824},
  volume={139},
  year={2021},
  publisher={PMLR}
}

@inproceedings{arnab2021vivit,
  title={{ViViT}: A Video Vision Transformer},
  author={Arnab, Anurag and Dehghani, Mostafa and Heigold, Georg and Sun, Chen and Lucic, Mario and Schmid, Cordelia},
  booktitle={Proceedings of the IEEE/CVF International Conference on Computer Vision (ICCV)},
  pages={6836--6846},
  year={2021},
  doi={10.1109/ICCV48922.2021.00676}
}

@inproceedings{fan2021multiscale,
  title={Multiscale Vision Transformers},
  author={Fan, Haoqi and Xiong, Bo and Mangalam, Karttikeya and Li, Yanghao and Yan, Zhicheng and Malik, Jitendra and Feichtenhofer, Christoph},
  booktitle={Proceedings of the IEEE/CVF International Conference on Computer Vision (ICCV)},
  pages={6824--6835},
  year={2021},
  doi={10.1109/ICCV48922.2021.00675}
}

@inproceedings{liu2022video,
  title={Video Swin Transformer},
  author={Liu, Ze and Ning, Jia and Cao, Yue and Wei, Yixuan and Zhang, Zheng and Lin, Stephen and Hu, Han},
  booktitle={Proceedings of the IEEE/CVF Conference on Computer Vision and Pattern Recognition (CVPR)},
  pages={3202--3211},
  year={2022},
  doi={10.1109/CVPR52688.2022.00320}
}

@inproceedings{li2022uniformer,
  title={{UniFormer}: Unified Transformer for Efficient Spatiotemporal Representation Learning},
  author={Li, Kunchang and Wang, Yali and Gao, Peng and Song, Guanglu and Liu, Yu and Li, Hongsheng and Qiao, Yu},
  booktitle={International Conference on Learning Representations (ICLR)},
  year={2022}
}

@inproceedings{tong2022videomae,
  title={{VideoMAE}: Masked Autoencoders are Data-Efficient Learners for Self-Supervised Video Pre-Training},
  author={Tong, Zhan and Song, Yibing and Wang, Jue and Wang, Limin},
  booktitle={Advances in Neural Information Processing Systems (NeurIPS)},
  volume={35},
  pages={10078--10093},
  publisher={Curran Associates, Inc.},
  year={2022}
}

@misc{wang2022internvideo,
  title={{InternVideo}: General Video Foundation Models via Generative and Discriminative Learning},
  author={Wang, Yi and Li, Kunchang and Li, Yizhuo and He, Yinan and Huang, Bingkun and Zhao, Zhiyu and Zhang, Hongjie and Xu, Jilan and Liu, Yi and Wang, Zun and Xing, Sen and Chen, Guo and Pan, Junting and Yu, Jiashuo and Wang, Yali and Wang, Limin and Qiao, Yu},
  year={2022},
  eprint={2212.03191},
  archivePrefix={arXiv},
  primaryClass={cs.CV},
  url={https://arxiv.org/abs/2212.03191}
}

@inproceedings{li2023uniformerv2,
  title={{UniFormerV2}: Unlocking the Potential of Image {ViT}s for Video Understanding},
  author={Li, Kunchang and Wang, Yali and He, Yinan and Li, Yizhuo and Wang, Yi and Wang, Limin and Qiao, Yu},
  booktitle={Proceedings of the IEEE/CVF International Conference on Computer Vision (ICCV)},
  pages={1632--1643},
  year={2023},
  doi={10.1109/ICCV51070.2023.00157}
}

@inproceedings{li2023unmasked,
  title={Unmasked Teacher: Towards Training-Efficient Video Foundation Models},
  author={Li, Kunchang and Wang, Yali and Li, Yizhuo and Wang, Yi and He, Yinan and Wang, Limin and Qiao, Yu},
  booktitle={Proceedings of the IEEE/CVF International Conference on Computer Vision (ICCV)},
  pages={19948--19960},
  year={2023},
  doi={10.1109/ICCV51070.2023.01826}
}

@inproceedings{wang2023videomaev2,
  title={{VideoMAE} V2: Scaling Video Masked Autoencoders with Dual Masking},
  author={Wang, Limin and Huang, Bingkun and Zhao, Zhiyu and Tong, Zhan and He, Yinan and Wang, Yi and Wang, Yali and Qiao, Yu},
  booktitle={Proceedings of the IEEE/CVF Conference on Computer Vision and Pattern Recognition (CVPR)},
  pages={14549--14560},
  year={2023},
  doi={10.1109/CVPR52729.2023.01398}
}

@misc{huang2023temporally,
  title={Temporally-Adaptive Models for Efficient Video Understanding},
  author={Huang, Ziyuan and Zhang, Shiwei and Pan, Liang and Qing, Zhiwu and Zhang, Yingya and Liu, Ziwei and Ang, Marcelo H.},
  year={2023},
  eprint={2308.05787},
  archivePrefix={arXiv},
  primaryClass={cs.CV},
  url={https://arxiv.org/abs/2308.05787}
}

@inproceedings{zhao2024asymmetric,
  title={Asymmetric Masked Distillation for Pre-Training Small Foundation Models},
  author={Zhao, Zhiyu and Huang, Bingkun and Xing, Sen and Wu, Gangshan and Qiao, Yu and Wang, Limin},
  booktitle={Proceedings of the IEEE/CVF Conference on Computer Vision and Pattern Recognition (CVPR)},
  pages={18516--18526},
  year={2024},
  doi={10.1109/CVPR52733.2024.01752}
}

@inproceedings{wang2024internvideo2,
  title={Internvideo2: Scaling foundation models for multimodal video understanding},
  author={Wang, Yi and Li, Kunchang and Li, Xinhao and Yu, Jiashuo and He, Yinan and Chen, Guo and Pei, Baoqi and Zheng, Rongkun and Wang, Zun and Shi, Yansong and others},
  booktitle={European conference on computer vision},
  pages={396--416},
  year={2024},
  organization={Springer}
}

@misc{wang2021actionclip,
  title={{ActionCLIP}: A New Paradigm for Video Action Recognition},
  author={Wang, Mengmeng and Xing, Jiazheng and Liu, Yong},
  year={2021},
  eprint={2109.08472},
  archivePrefix={arXiv},
  primaryClass={cs.CV},
  url={https://arxiv.org/abs/2109.08472}
}

@inproceedings{moosavi2017universal,
  title={Universal Adversarial Perturbations},
  author={Moosavi-Dezfooli, Seyed-Mohsen and Fawzi, Alhussein and Fawzi, Omar and Frossard, Pascal},
  booktitle={Proceedings of the IEEE Conference on Computer Vision and Pattern Recognition (CVPR)},
  pages={1765--1773},
  year={2017},
  doi={10.1109/CVPR.2017.17}
}

@inproceedings{kurakin2017adversarialscale,
  title={Adversarial Machine Learning at Scale},
  author={Kurakin, Alexey and Goodfellow, Ian and Bengio, Samy},
  booktitle={International Conference on Learning Representations (ICLR)},
  year={2017}
}

@inproceedings{dong2018boosting,
  title={Boosting Adversarial Attacks with Momentum},
  author={Dong, Yinpeng and Liao, Fangzhou and Pang, Tianyu and Su, Hang and Zhu, Jun and Hu, Xiaolin and Li, Jianguo},
  booktitle={Proceedings of the IEEE/CVF Conference on Computer Vision and Pattern Recognition (CVPR)},
  pages={9185--9193},
  year={2018},
  doi={10.1109/CVPR.2018.00957}
}

@inproceedings{xiao2018spatially,
  title={Spatially Transformed Adversarial Examples},
  author={Xiao, Chaowei and Zhu, Jun-Yan and Li, Bo and He, Warren and Liu, Mingyan and Song, Dawn},
  booktitle={International Conference on Learning Representations (ICLR)},
  year={2018}
}

@inproceedings{andriushchenko2020square,
  title={Square Attack: A Query-Efficient Black-Box Adversarial Attack via Random Search},
  author={Andriushchenko, Maksym and Croce, Francesco and Flammarion, Nicolas and Hein, Matthias},
  booktitle={Proceedings of the European Conference on Computer Vision (ECCV)},
  pages={484--501},
  year={2020},
  publisher={Springer}
}

@article{li2021adversarial,
  title={Adversarial attacks on black box video classifiers: Leveraging the power of geometric transformations},
  author={Li, Shasha and Aich, Abhishek and Zhu, Shitong and Asif, Salman and Song, Chengyu and Roy-Chowdhury, Amit and Krishnamurthy, Srikanth},
  journal={Advances in Neural Information Processing Systems},
  volume={34},
  pages={2085--2096},
  year={2021}
}

@inproceedings{luo2022frequency,
  title={Frequency-driven Imperceptible Adversarial Attack on Semantic Similarity},
  author={Luo, Cheng and Lin, Qinliang and Xie, Weicheng and Wu, Bizhu and Xie, Jinheng and Shen, Linlin},
  booktitle={Proceedings of the IEEE/CVF Conference on Computer Vision and Pattern Recognition (CVPR)},
  pages={15294--15303},
  year={2022},
  doi={10.1109/CVPR52688.2022.01488}
}

@inproceedings{cao2023stylefool,
  title={{StyleFool}: Fooling Video Classification Systems via Style Transfer},
  author={Cao, Yuxin and Xiao, Xi and Sun, Ruoxi and Wang, Derui and Xue, Minhui and Wen, Sheng},
  booktitle={2023 IEEE Symposium on Security and Privacy (SP)},
  pages={1631--1648},
  year={2023},
  publisher={IEEE},
  doi={10.1109/SP46215.2023.10179383}
}

@inproceedings{li2025bmtc,
  title={Temporal Consistency Constrained Transferable Adversarial Attacks with Background Mixup for Action Recognition},
  author={Li, Ping and Ni, Jianan and Pang, Bo},
  booktitle={Proceedings of the Thirty-Fourth International Joint Conference on Artificial Intelligence (IJCAI)},
  pages={1413--1421},
  year={2025},
  doi={10.24963/ijcai.2025/158},
  url={https://www.ijcai.org/proceedings/2025/158}
}

@inproceedings{korhonen2022adversarial,
  title={Adversarial Attacks Against Blind Image Quality Assessment Models},
  author={Korhonen, Jari and You, Junyong},
  booktitle={Proceedings of the 2nd Workshop on Quality of Experience in Visual Multimedia Applications (QoEVMA '22)},
  pages={3--11},
  year={2022},
  publisher={ACM},
  doi={10.1145/3552469.3555715}
}

@inproceedings{zhang2022perceptual,
  title={Perceptual Attacks of No-Reference Image Quality Models with Human-in-the-Loop},
  author={Zhang, Weixia and Li, Dingquan and Min, Xiongkuo and Zhai, Guangtao and Guo, Guodong and Yang, Xiaokang and Ma, Kede},
  booktitle={Advances in Neural Information Processing Systems (NeurIPS)},
  volume={35},
  pages={2916--2929},
  year={2022},
  publisher={Curran Associates, Inc.}
}

@inproceedings{guo2018countering,
  title={Countering Adversarial Images using Input Transformations},
  author={Guo, Chuan and Rana, Mayank and Cisse, Moustapha and van der Maaten, Laurens},
  booktitle={International Conference on Learning Representations (ICLR)},
  year={2018}
}

@inproceedings{xie2018mitigating,
  title={Mitigating Adversarial Effects Through Randomization},
  author={Xie, Cihang and Wang, Jianyu and Zhang, Zhishuai and Ren, Zhou and Yuille, Alan},
  booktitle={International Conference on Learning Representations (ICLR)},
  year={2018}
}

@inproceedings{cohen2019certified,
  title={Certified Adversarial Robustness via Randomized Smoothing},
  author={Cohen, Jeremy M. and Rosenfeld, Elan and Kolter, J. Zico},
  booktitle={Proceedings of the 36th International Conference on Machine Learning (ICML)},
  pages={1310--1320},
  volume={97},
  year={2019},
  publisher={PMLR}
}

@article{hwang2024temporal,
  title={Temporal shuffling for defending deep action recognition models against adversarial attacks},
  author={Hwang, Jaehui and Zhang, Huan and Choi, Jun-Ho and Hsieh, Cho-Jui and Lee, Jong-Seok},
  journal={Neural Networks},
  volume={169},
  pages={388--397},
  year={2024},
  publisher={Elsevier}
}

@article{jiang2025videopure,
  title={Videopure: Diffusion-based adversarial purification for video recognition},
  author={Jiang, Kaixun and Chen, Zhaoyu and Fu, Jiyuan and Hong, Lingyi and Li, Jinglun and Zhang, Wenqiang},
  journal={IEEE Transactions on Circuits and Systems for Video Technology},
  year={2025},
  publisher={IEEE}
}

@article{pei2025freqpure,
  title={Diffusion-based adversarial purification from the perspective of the frequency domain},
  author={Pei, Gaozheng and Ma, Ke and Sun, Yingfei and Xu, Qianqian and Huang, Qingming},
  journal={arXiv preprint arXiv:2505.01267},
  year={2025}
}

@techreport{soomro2012ucf101,
  title={{UCF101}: A Dataset of 101 Human Actions Classes From Videos in The Wild},
  author={Soomro, Khurram and Zamir, Amir Roshan and Shah, Mubarak},
  institution={Center for Research in Computer Vision, University of Central Florida},
  number={CRCV-TR-12-01},
  year={2012}
}

@misc{kay2017kinetics,
  title={The Kinetics Human Action Video Dataset},
  author={Kay, Will and Carreira, Joao and Simonyan, Karen and Zhang, Brian and Hillier, Chloe and Vijayanarasimhan, Sudheendra and Viola, Fabio and Green, Tim and Back, Trevor and Natsev, Paul and Suleyman, Mustafa and Zisserman, Andrew},
  year={2017},
  eprint={1705.06950},
  archivePrefix={arXiv},
  primaryClass={cs.CV},
  url={https://arxiv.org/abs/1705.06950}
}

@inproceedings{goyal2017something,
  title={The ``Something Something'' Video Database for Learning and Evaluating Visual Common Sense},
  author={Goyal, Raghav and Ebrahimi Kahou, Samira and Michalski, Vincent and Materzynska, Joanna and Westphal, Susanne and Kim, Heuna and Haenel, Valentin and Fruend, Ingo and Yianilos, Peter and Mueller-Freitag, Moritz and Hoppe, Florian and Thurau, Christian and Bax, Ingo and Memisevic, Roland},
  booktitle={Proceedings of the IEEE International Conference on Computer Vision (ICCV)},
  pages={5842--5850},
  year={2017},
  doi={10.1109/ICCV.2017.622}
}

@inproceedings{zhang2018unreasonable,
  title={The Unreasonable Effectiveness of Deep Features as a Perceptual Metric},
  author={Zhang, Richard and Isola, Phillip and Efros, Alexei A. and Shechtman, Eli and Wang, Oliver},
  booktitle={Proceedings of the IEEE/CVF Conference on Computer Vision and Pattern Recognition (CVPR)},
  pages={586--595},
  year={2018},
  doi={10.1109/CVPR.2018.00068}
}

@ARTICLE{wang2004image,
  author={Wang, Zhou and Bovik, Alan C. and Sheikh, Hamid R. and Simoncelli, Eero P.},
  journal={IEEE Transactions on Image Processing},
  title={Image Quality Assessment: From Error Visibility to Structural Similarity},
  year={2004},
  volume={13},
  number={4},
  pages={600--612},
  doi={10.1109/TIP.2003.819861}
}

@ARTICLE{ding2022dists,
  author={Ding, Keyan and Ma, Kede and Wang, Shiqi and Simoncelli, Eero P.},
  journal={IEEE Transactions on Pattern Analysis and Machine Intelligence},
  title={Image Quality Assessment: Unifying Structure and Texture Similarity},
  year={2022},
  volume={44},
  number={5},
  pages={2567--2581},
  doi={10.1109/TPAMI.2020.3045810}
}

@misc{2020mmaction2,
    title={OpenMMLab's Next Generation Video Understanding Toolbox and Benchmark},
    author={MMAction2 Contributors},
    howpublished = {\url{https://github.com/open-mmlab/mmaction2}},
    year={2020}
}

@inproceedings{fan2021pytorchvideo,
  title={{PyTorchVideo}: A Deep Learning Library for Video Understanding},
  author={Fan, Haoqi and Murrell, Tullie and Wang, Heng and Alwala, Kalyan Vasudev and Li, Yanghao and Li, Yilei and Xiong, Bo and Ravi, Nikhila and Li, Meng and Yang, Haichuan and Malik, Jitendra and Girshick, Ross and Feiszli, Matt and Adcock, Aaron and Lo, Wan-Yen and Feichtenhofer, Christoph},
  booktitle={Proceedings of the 29th ACM International Conference on Multimedia},
  pages={3783--3786},
  year={2021},
  publisher={ACM},
  doi={10.1145/3474085.3478329}
}

@inproceedings{radford2021learning,
  title={Learning Transferable Visual Models From Natural Language Supervision},
  author={Radford, Alec and Kim, Jong Wook and Hallacy, Chris and Ramesh, Aditya and Goh, Gabriel and Agarwal, Sandhini and Sastry, Girish and Askell, Amanda and Mishkin, Pamela and Clark, Jack and Krueger, Gretchen and Sutskever, Ilya},
  booktitle={Proceedings of the 38th International Conference on Machine Learning (ICML)},
  pages={8748--8763},
  volume={139},
  year={2021},
  publisher={PMLR}
}

@inproceedings{dosovitskiy2021image,
  title={An Image is Worth 16x16 Words: Transformers for Image Recognition at Scale},
  author={Dosovitskiy, Alexey and Beyer, Lucas and Kolesnikov, Alexander and Weissenborn, Dirk and Zhai, Xiaohua and Unterthiner, Thomas and Dehghani, Mostafa and Minderer, Matthias and Heigold, Georg and Gelly, Sylvain and Uszkoreit, Jakob and Houlsby, Neil},
  booktitle={International Conference on Learning Representations (ICLR)},
  year={2021}
}

@article{papernot2016technical,
  title={Technical report on the cleverhans v2. 1.0 adversarial examples library},
  author={Papernot, Nicolas and Faghri, Fartash and Carlini, Nicholas and Goodfellow, Ian and Feinman, Reuben and Kurakin, Alexey and Xie, Cihang and Sharma, Yash and Brown, Tom and Roy, Aurko and others},
  journal={arXiv preprint arXiv:1610.00768},
  year={2016}
}

@misc{rauber2017foolbox,
  title         = {Foolbox: A Python toolbox to benchmark the robustness of machine learning models},
  author        = {Jonas Rauber and Wieland Brendel and Matthias Bethge},
  year          = {2017},
  eprint        = {1707.04131},
  archivePrefix = {arXiv},
  primaryClass  = {cs.LG},
  doi           = {10.48550/arXiv.1707.04131}
}

@inproceedings{croce2020reliable,
  title={Reliable evaluation of adversarial robustness with an ensemble of diverse parameter-free attacks},
  author={Croce, Francesco and Hein, Matthias},
  booktitle={International conference on machine learning},
  pages={2206--2216},
  year={2020},
  organization={PMLR}
}

@inproceedings{croce2021robustbench,
  title     = {RobustBench: a standardized adversarial robustness benchmark},
  author    = {Francesco Croce and Maksym Andriushchenko and Vikash Sehwag and Edoardo Debenedetti and Nicolas Flammarion and Mung Chiang and Prateek Mittal and Matthias Hein},
  booktitle = {Thirty-fifth Conference on Neural Information Processing Systems Datasets and Benchmarks Track},
  year      = {2021}
}

@inproceedings{tu2023ila,
  author    = {Shuyuan Tu and Qi Dai and Zuxuan Wu and Zhi-Qi Cheng and Han Hu and Yu-Gang Jiang},
  title     = {Implicit Temporal Modeling with Learnable Alignment for Video Recognition},
  booktitle = {Proceedings of the IEEE/CVF International Conference on Computer Vision (ICCV)},
  pages     = {19936--19947},
  year      = {2023}
}

@misc{wang2024onepeace,
  title={{ONE-PEACE}: Exploring One General Representation Model Toward Unlimited Modalities},
  author={Wang, Peng and Wang, Shijie and Lin, Junyang and Bai, Shuai and Zhou, Xiaohuan and Zhou, Jingren and Wang, Xinggang and Zhou, Chang},
  year={2023},
  eprint={2305.11172},
  archivePrefix={arXiv},
  primaryClass={cs.CV},
  url={https://arxiv.org/abs/2305.11172}
}

@inproceedings{liu2021tam,
  title={{TAM}: Temporal Adaptive Module for Video Recognition},
  author={Liu, Zhaoyang and Wang, Limin and Wu, Wayne and Qian, Chen and Lu, Tong},
  booktitle={Proceedings of the IEEE/CVF International Conference on Computer Vision (ICCV)},
  pages={13708--13718},
  year={2021},
  doi={10.1109/ICCV48922.2021.01345}
}

@inproceedings{li2022mvitv2,
  title={{MViTv2}: Improved Multiscale Vision Transformers for Classification and Detection},
  author={Li, Yanghao and Wu, Chao-Yuan and Fan, Haoqi and Mangalam, Karttikeya and Xiong, Bo and Malik, Jitendra and Feichtenhofer, Christoph},
  booktitle={Proceedings of the IEEE/CVF Conference on Computer Vision and Pattern Recognition (CVPR)},
  pages={4794--4804},
  year={2022},
  doi={10.1109/CVPR52688.2022.00476}
}

\end{document}